\documentclass[runningheads]{llncs}
\usepackage[T1]{fontenc}
\usepackage{graphicx}
\usepackage{outlines}
\usepackage{xcolor}
\usepackage{float}
\usepackage{caption}
\usepackage{amssymb}
\usepackage{amsmath}
\usepackage{hyperref}
\usepackage{marvosym}

\begin{document}
\title{SpaRED benchmark: Enhancing Gene Expression Prediction from Histology Images with Spatial Transcriptomics Completion}
% Benchmarking Gene Expression Completion \& Prediction for Spatial Transcriptomics
% Transcriptomic Completion Improves Expression Prediction from Histology Images in a Comprehensive Benchmark
%Gene Expression Completion with Transformers for Spatial Transcriptomics
%
\titlerunning{SpaRED}
\author{Gabriel Mejia$^*$\Letter
\and
Daniela Ruiz$^*$
\and
Paula Cárdenas
\and
Leonardo Manrique
\and
Daniela Vega
\and
Pablo Arbeláez}
%index{Mejia, Gabriel}
%index{Ruiz, Daniela}
%index{Cárdenas, Paula}
%index{Manrique, Leonardo}
%index{Vega, Daniela}
%index{Arbelaez, Pablo}

\authorrunning{G. Mejia, et al.}
\institute{Center for Research and Formation in Artificial Intelligence \\ Universidad de los Andes, Colombia \\
\email{\{gm.mejia,da.ruizl1,p.cardenasg,dl.manrique,d.vegaa\}@uniandes.edu.co}}
\maketitle              % typeset the header of the contribution

\begin{abstract}

Spatial Transcriptomics is a novel technology that aligns histology images with spatially resolved gene expression profiles. Although groundbreaking, it struggles with gene capture yielding high corruption in acquired data. Given potential applications, recent efforts have focused on predicting transcriptomic profiles solely from histology images. However, differences in databases, preprocessing techniques, and training hyperparameters hinder a fair comparison between methods. To address these challenges, we present a systematically curated and processed database collected from 26 public sources, representing an 8.6-fold increase compared to previous works. Additionally, we propose a state-of-the-art transformer-based completion technique for inferring missing gene expression, which significantly boosts the performance of transcriptomic profile predictions across all datasets. Altogether, our contributions constitute the most comprehensive benchmark of gene expression prediction from histology images to date and a stepping stone for future research on spatial transcriptomics.

\end{abstract}
\section{Introduction}

Spatial Transcriptomics (ST) is an emerging technology that precisely localizes gene expression profiles within histological images \cite{jiang2023generalization}. While histology analysis is the gold standard for diagnosis of many diseases \cite{xie2023spatially}, transcriptomics unlocks molecular insights that unveil causal pathways behind pathologies \cite{zeng2022spatial,jiang2023generalization}. Together, they open a new spectrum of possibilities to understand how physiological alterations start, evolve and respond to treatment directly in spatial coordinates \cite{wang2023crost}. 

Although potentially groundbreaking, the deployment of ST presents significant practical challenges. On the technical side, it inherits the problem of not detecting transcripts even though they were present in the source tissue, a failure mode known as dropout in bulk and single-cell transcriptomics \cite{pham2023robust,avsar2023comparative}. In practice, this phenomenon is observed as pepper noise in the acquired gene expression maps and is so severe that single-cell reference datasets are commonly required to handle missing data \cite{avsar2023comparative}.

\begin{figure}[t]
    \includegraphics[width=0.99\textwidth]{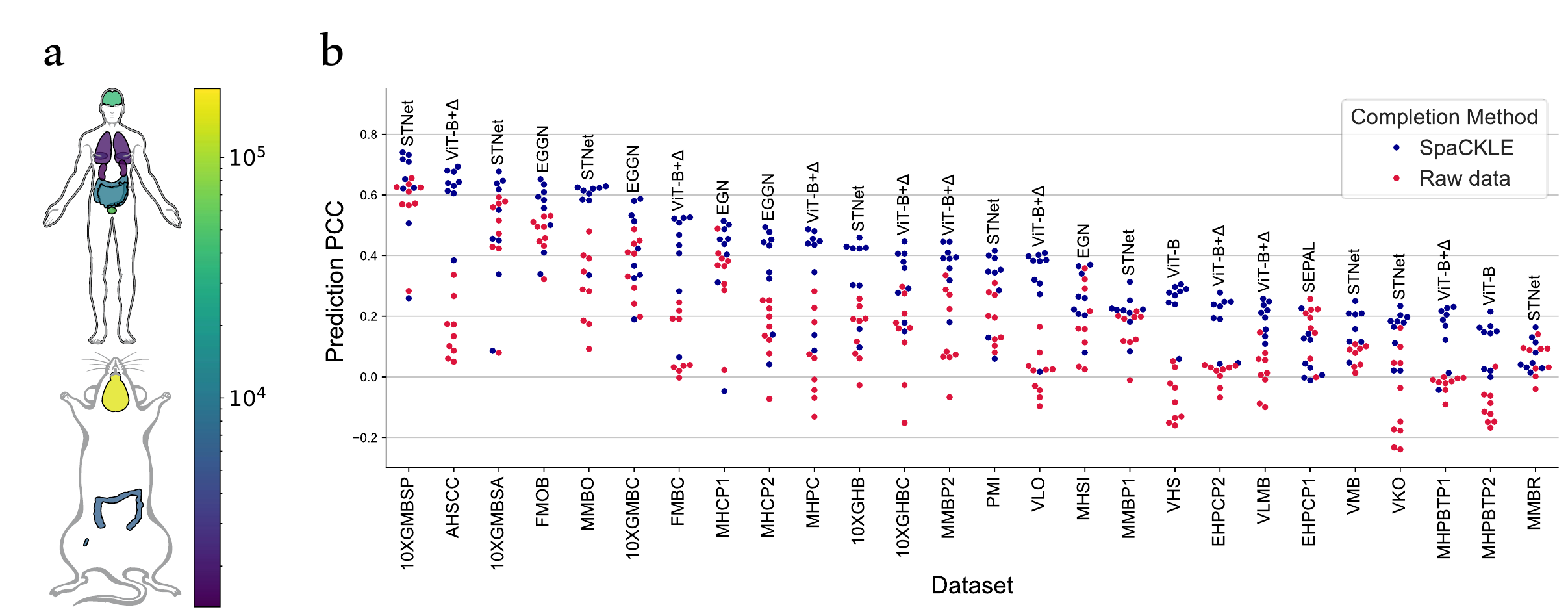}
    \caption{(a) Organisms and tissues available in SpaRED, along with the number of spots available from each tissue. (b) Prediction Pearson Correlation Coefficient for each model across all the datasets in SpaRED. For each dataset, the state-of-the-art model that obtains the highest Pearson Correlation Coefficient is included.}
    \label{fig:general_results}
\end{figure}

Among the social challenges, expensive equipment, the need for domain expertise, and a slow adoption in clinical settings impede diagnostic benefits from reaching most patients \cite{pang2021leveraging}. Acknowledging these problems, the deep learning community has delved into democratizing ST by studying gene expression prediction from histology images \cite{jiang2023generalization}. Solving this problem would allow patients to obtain all the molecular insights from the image of an ordinary biopsy.

As with any novel technology, multiple variations of ST are currently in use \cite{slide_seq_v2,merfish,10x_visium}. However, as demonstrated by the number of entries in the comprehensive repository of spatial transcriptomics \cite{wang2023crost}, 10X Visium \cite{10x_visium} is by far the most popular. Leveraging the abundance of public Visium data, several deep learning methods have been developed for this task \cite{he2020integrating,pang2021leveraging,yang2023exemplar,yang2024spatial,xie2023spatially,zeng2022spatial,mejia2023SEPAL}, varying greatly in architecture and always reporting favorable results against state-of-the-art. Nonetheless, differences in databases, data preprocessing, and training hyperparameters impede fair comparison between approaches and compromise the validity of new results.

To address these challenges, we present two contributions. First, we systematically compile, curate and standardize 26 public ST databases into the \textbf{Spa}tially \textbf{R}esolved \textbf{E}xpression \textbf{D}atabase (SpaRED), an extensive Visium resource with human and mouse samples from 9 tissue types as shown in Fig. \ref{fig:general_results}.a. Secondly, we evaluate 7 state-of-the-art prediction methods in SpaRED, establishing a new benchmark with an 8.6-fold increase in analyzed datasets compared to previous works \cite{jiang2023generalization}. 

As main technical contribution, inspired by the unrivaled power of self-attention mechanisms for next token prediction in natural language processing \cite{dosovitskiy2020image}, we design a transformer-based completion model for corrupted gene expression vectors, which we call \textbf{Spa}tial transcriptomics \textbf{C}ompletion with \textbf{K}nowledge from the \textbf{L}ocal \textbf{E}nvironment (SpaCKLE). SpaCKLE not only outperforms previous gene completion strategies but significantly enhances the gene prediction performance of all state-of-the-art methods across every dataset of our benchmark, as evidenced by the improvement from red to blue points in Fig.\ref{fig:general_results}b. Our project’s benchmark and source code is publicly available at \\ \href{https://bcv-uniandes.github.io/spared_webpage/}{https://bcv-uniandes.github.io/spared\_webpage/}.

\section{Related Work}

\subsection{Integrated Databases}

Recent advancements in ST have led to the development of multiple databases. For instance, CROST \cite{wang2023crost} is a comprehensive repository with 1033 spatial transcriptomics samples from 8 species, 35 tissues, and 56 diseases. Other databases include SpatialDB, Aquila, SPASCER, SODB, and STomicsDB \cite{wang2023crost}, each offering unique datasets and analytical tools. Although these databases facilitate advanced spatial analyses, they are not designed for the expression profile prediction task. SpaRED tackles this limitation by implementing best practices in bioinformatics analysis, including the selection of Moran genes, standardization of reference genomes, TPM normalization, and batch correction, making SpaRED particularly valuable for clinical applications. 

\subsection{Completion strategies}

To address the missing value problem, \cite{mejia2023SEPAL} employs a modified adaptive median filter as a completion strategy, replacing dropout values with circular region medians or, if unsuccessful, the whole slide image (WSI)'s median. Alternatively, stLearn \cite{pham2023robust} uses genetic and morphological similarity to adjust existing spots or predict gene expression for missing values. Moreover, although there are alternative methods that integrate knowledge from single cell RNAseq (Seurat \cite{stuart2019comprehensive}, Harmony \cite{korsunsky2019fast}, LIGER \cite{welch2019single} and Tangram \cite{biancalani2021deep}), these strategies require a paired single-cell dataset that hampers their usability and practicality. In contrast, SpaCKLE is a reference-free completion method, which stands out from alternatives by leveraging the complete genetic profile of adjacent spots and taking advantage of the transformer capacity to predict missing values.

\subsection{Gene Expression Prediction Benchmarks}

A recent study by \cite{jiang2023generalization} reviews 6 deep learning methods for gene expression profile prediction, testing their performance on three distinct breast cancer datasets. Although the study presents a solid model performance analysis, it only focuses on human breast cancer tissue. Hence, our benchmark represents a substantial advancement, featuring an 8.6-fold increase in the number of datasets and including 9 different tissue types from human and mouse subjects.

\section{Spatially Resolved Expression Database}
 
\subsection{Original Datasets and Curation} 

To build SpaRED, we collect raw data from 7 independent publications \cite{Abalo2021}, \cite{parigi2022spatial}, \cite{villacampa2021genome}, \cite{vicari2023spatial}, \cite{mirzazadeh2023spatially}, \cite{erickson2022spatially}, \cite{fan2023expansion}
and complement them using 5 demonstration datasets from 10X Genomics (available through the SquidPy python package \cite{palla2022squidpy}). We only include datasets with more than one WSI and split the publications' data by tissue type, resulting in 26 distinct datasets: 14 from human and 12 from mouse, showcasing a variety of tissue samples, as illustrated in Fig. \ref{fig:general_results}.a. According to the number of patients in a dataset, we define two types of tasks: intra-patient generalization (where WSIs are generally consecutive cuts of a single tissue) and inter-patient generalization (where WSIs correspond to the same tissue in different subjects). We manually assign WSIs into train, validation, or test sets seeking similar visual distributions in all splits. In 11 out of 26 cases, a test set is defined. Otherwise, due to the limited number of patients/slides, we split the data into train and validation sets.

For data preprocessing, we follow the protocol proposed in \cite{mejia2023SEPAL}, which handles batch correction, normalization, and gene selection based on Moran's I calculations. Depending on data quality, we select either 32 (6/26 datasets) or 128 (20/26 datasets) prediction genes in each refined dataset. In total, SpaRED contains 105 slides (308,843 spots) collected from 35 patients. See supplementary Table 1 for detailed datasets statistics.

\subsection{Benchmark of Existing Gene Prediction Methods}

We use SpaRED to evaluate 7 state-of-the-art expression profile prediction methods. Among these, HisToGene \cite{pang2021leveraging} splits a WSI into patches that are processed by a Visual Transformer (ViT) model. The output is the genetic profile of the WSI. STNet \cite{he2020integrating} inputs individual patches into a fine-tuned DenseNet-121 with a linear layer for prediction. Additionally, STNet averages predictions across 8 symmetries of each patch to determine the final output. EGN \cite{yang2023exemplar} and its improved version EGGN \cite{yang2024spatial} apply exemplar-guided learning, a prediction strategy that bases its estimations on patches that are visually similar to the target patch within a latent space. BLEEP \cite{xie2023spatially}, employs contrastive learning to map image patches and expression profiles in a shared latent space. Hist2ST \cite{zeng2022spatial} applies a Convolutional Neural Network (CNN) to extract local patch features, followed by a transformer and a Graph Neural Network (GNN) to handle long and near-range dependencies, respectively. Finally, SEPAL\cite{mejia2023SEPAL} finetunes a ViT backbone, and subsequently refines its predictions applying a GNN that processes a neighborhood graph for each patch. Additionally, SEPAL supervises expression changes relative to the mean expression in the training data instead of the absolute expression value, a strategy denoted ($\Delta$) prediction. 

Alongside these models, our comprehensive benchmark also includes the performance of three baseline methods: a ShuffleNet \cite{zhang2018shufflenet} architecture that finetunes an image encoder with low computational cost, a ViT-B encoder \cite{dosovitskiy2020image} that reflects the impact of fine-tuning a state-of-the-art backbone for this task, and a ViT-B+$\Delta$ approach as suggested by \cite{mejia2023SEPAL}.

We search for the optimal learning rate in every dataset, and then, with this value fixed, we explore two training scenarios: using raw data directly and SpaCKLE-completed data.

\begin{figure}[t]  
    \includegraphics[width=0.99\textwidth]{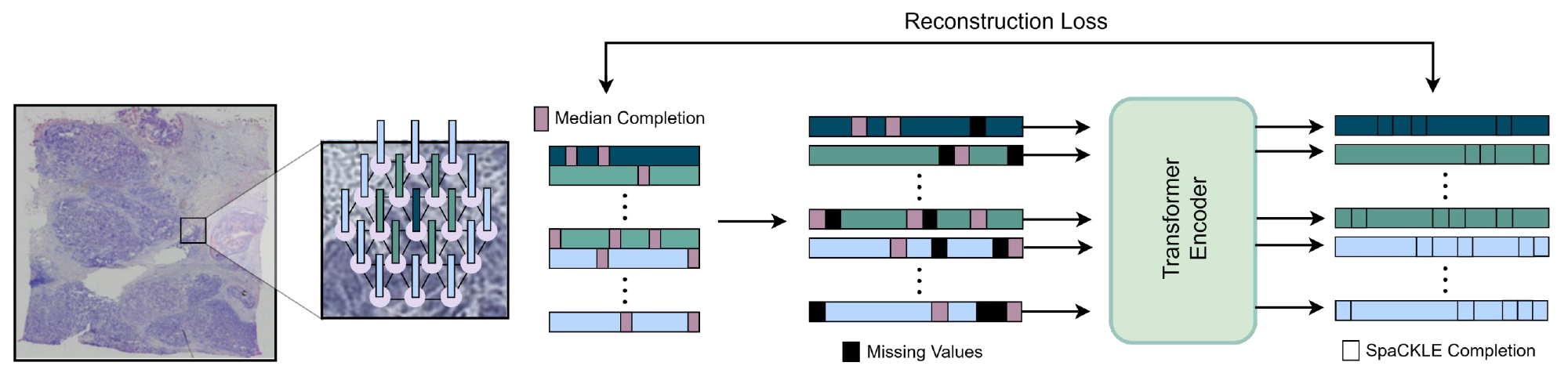}
    \caption{Overview of our data completion framework using a transformer-based model.}
    \label{fig:imputation}
\end{figure}

\section{Gene Completion with Transformers}

%Upon the processing phase, we initially completed data following the imputation mechanism proposed in \cite{mejia2023SEPAL} to train SpaCKLE. As shown in Figure \ref{fig:imputation}A, we extract data from a 2-hop vicinity of each patch and mask it randomly with a 30\% probability. This mask does not overlap with previously imputed values to avoid using synthetic data as ground truths. We then process the raw data with 2 transformer encoder layers \cite{vaswani2023attention} and apply an MSE reconstruction loss between the input and the output sequence. Although the model uses all values (including median imputed ones) for supervision during training, we only compute metrics over the randomly masked set of the first token. This approach ensures we test the model's completion ability in known data. At completion time, we use the original missing data as a random mask and fill in missing values using the first token of the output sequence.

%SpaCKLE consists of a transformer-based architecture designed to complete gene expression data where the ST technology failed to measure transcripts.

Inspired by the disruptive success of the transformer architecture for completion tasks such as language next token prediction \cite{vaswani2023attention} and visual reconstruction \cite{He2021}, we adapt these ideas to the ST domain. Fig. \ref{fig:imputation} illustrates SpaCKLE's training, a process that takes as a starting point data that we pre-completed using the median method proposed in \cite{mejia2023SEPAL}. This process ensures faster training convergence and guarantees non-zero predictions. Given the median-completed expression vector $x \in \mathbb{R}^{[g,1]}$ of a particular spot with $g$ prediction genes, we start by extracting the expression matrix $E_x = [x; V_x] \in \mathbb{R}^{[g,n+1]}$, which concatenates $x$ with the expression matrix $V_x \in \mathbb{R}^{[g, n]}$ of its $n$ 2-hop neighbors in the Visium hexagonal geometry. We then randomly mask $E_x$ and process it with a transformer encoder $T(\cdot)$ that leverages the self-attention mechanism:
\begin{align}
    \text{Attention}(Q, K, V) = \text{softmax}\left(\frac{QK^T}{\sqrt{d_k}}\right) V, 
\end{align}
to get a reconstructed version $\hat{E}_x$ as follows:
\begin{align}
    E_m &= E_x \odot M(\rho)\\
    \hat{E}_x &= L_{out}\left(T(L_{in}(E_m))\right),
\end{align}

where $\odot$ represents the Hadamard product between $E_x$ and a binary masking matrix $M \in \mathbb{R}^{[g, n+1]}$ that does not overlap with median-completed values and exhibits a zero value probability $\rho=30\%$. To accommodate different gene dimensionalities to a fixed transformer dimension $d_k=128$, we use the $L_{in}(\cdot)$ and $L_{out}(\cdot)$ linear adapters. We optimize a Mean Square Error (MSE) loss between the two complete matrices $\mathcal{L}=\left \| E_x - \hat{E}_x \right \|_2$ but we only compute metrics and complete missing values using the masked elements from the first vector of the output. Hence, each component of the completed version $\hat{x}$ can be expressed as:
\begin{align}
    \hat{x}_i = 
    \begin{cases} 
      x_i, & M[1, i]=1 \\
      \hat{E}_x[1, i], & M[1, i]=0 
   \end{cases}
\end{align}
At inference, we replace $M$ with a binary matrix indicating original missing values to get a refined version of the gene expression profiles.

\begin{figure}[t]  
    \includegraphics[width=0.99\textwidth]{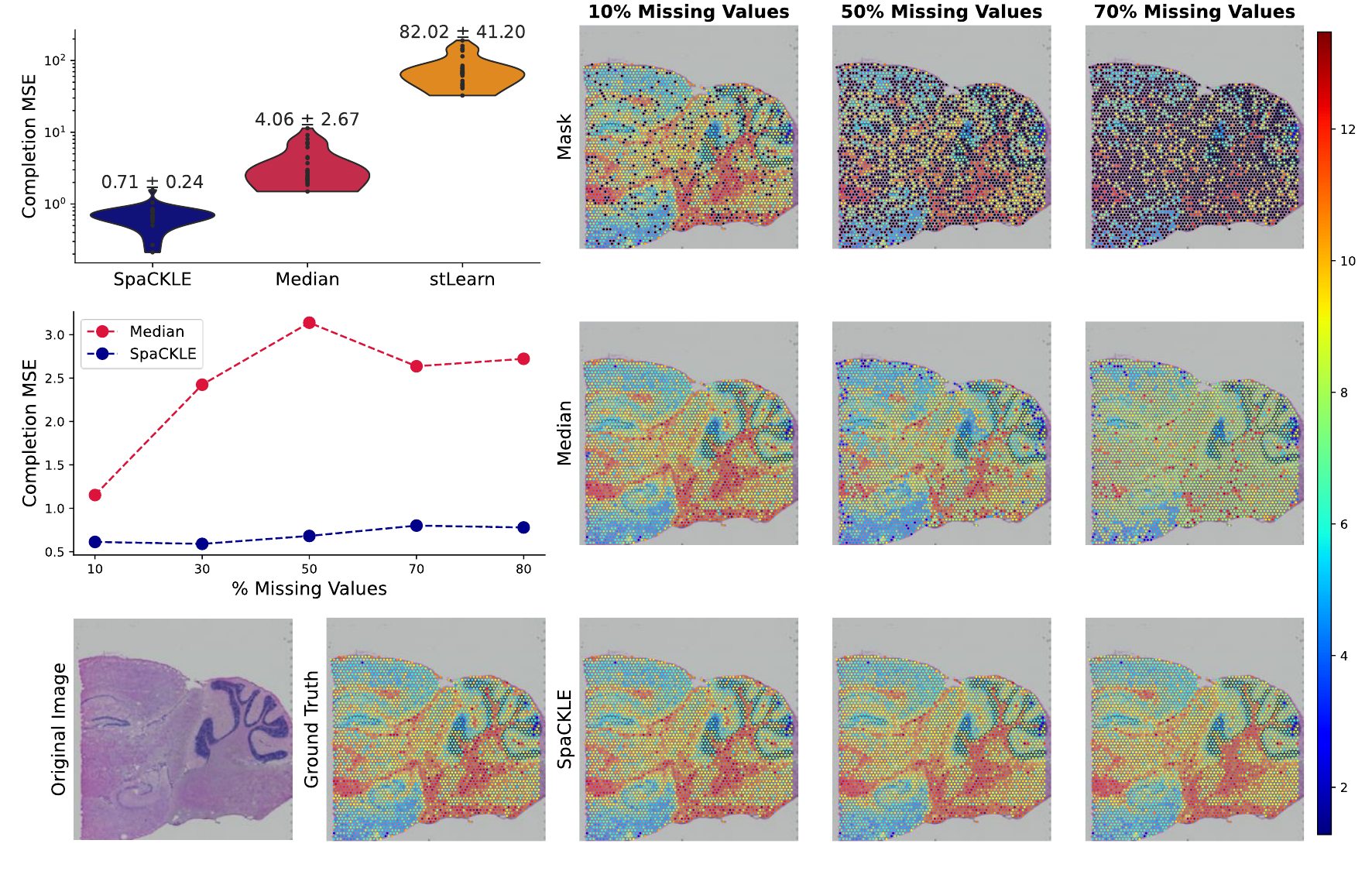}
    \caption{Completion results: Violin plot displaying completion MSE scores for each method (SpaCKLE, Median and stLearn) across all datasets in SpaRED (upper left). Line plot displaying completion MSE for the median and SpaCKLE methods across different percentages of synthetically masked data (middle left). Qualitative results showing gene completion for increasing synthetic masking percentages (row 1) with the median method (row 2) and SpaCKLE (row 3).}
    \label{fig:imputation_2}
\end{figure}

\subsubsection{Implementation Details:} We train all our models on a NVIDIA Quadro RTX 8000 with a batch size of 256 and use an Adam \cite{kingma2017adam} optimizer with default PyTorch library parameters. We handle both regression and completion problems as multivariate regression tasks and evaluate them using MSE and Pearson Correlation Coefficient (PCC). To select the best model, we save the one with the lowest validation MSE after 1,000 and 10,000 iterations for prediction and completion, respectively. All metrics are computed exclusively on real data for both the completion and the prediction task.

\section{Results and Discussion}

\subsection{Gene Completion Evaluation}

The violin plot in Fig. \ref{fig:imputation_2} presents a comparison of the logarithmic MSE for data completion using SpaCKLE, the median completion method, and stLearn across SpaRED. The results indicate that SpaCKLE outperforms alternative completion methods, with a relative 82.5\% MSE reduction compared to the median method and by two orders of magnitude concerning stLearn. Notably, stLearn presents the highest MSE in the entirety of SpaRED, which conveys its inability to restore masked data. These results are consistent with those reported in \cite{avsar2023comparative}, where stLearn's completion predictions included a high proportion of zero values. It is noteworthy that the median method is based solely on the adjacent expression of a single gene, an approach that, although straightforward, does not consider the broader genetic context. In contrast, SpaCKLE has access to the complete genetic profile of the neighboring spots. Thus, we hypothesize that our transformer architecture is leveraging the full expression profile of the empty spot's vicinity to enhance completion predictions.

To thoroughly assess the robustness of our approach, we characterize the completion performance when synthetically corrupting increasing percentages of data in the 10XGMBSP dataset. The MSE results in Fig. \ref{fig:imputation_2} show how the completion's accuracy changes with various masking percentages. For visualization purposes, we only show SpaCKLE and the median method since stLearn has a significantly higher MSE. We observe that, as the task gets more challenging with a greater percentage of missing data, SpaCKLE outperforms the median completion method by a larger margin. The predicted expression maps support these observations, showing that SpaCKLE strongly approximates the ground truth patterns even at a missing value percentage of 70\%. Conversely, the uniformity in the color pattern of the predictions from the median method demonstrates that this strategy repeatedly imposes the global median when it cannot find a local value due to the high fraction of missing data. This behavior impairs the expression profiles by homogenizing the gene's activity in the tissue and removing spatial information. 

% focus on the ability of SpaCKLE and of the median approach to complete several levels of missing data. Our strategy consists of incrementally and randomly masking known expression values in the dataset, thereby simulating different data corruption scenarios. To assess each method's performance, we calculate the MSE between the predictions and the real expression only considering the masked elements. 

%The MSE results in Fig. \ref{fig:imputation_2} show how the completion's accuracy changes with various masking percentages. Naturally, when a larger proportion of data is missing, the completion task becomes more challenging for both methods. Nevertheless, SpaCKLE outperforms the median completion method across varying percentages of missing data. In addition, as the fraction of missing values increases, the superiority of SpaCKLE becomes more evident. The predicted expression maps support these observations, showing that SpaCKLE strongly approximates the ground truth patterns even at missing value percentages of 80\%. On the contrary, the uniformity in the color pattern of the predictions from the medians demonstrates that this method repeatedly imposes the global median when it cannot find a local value due to the high fraction of missing data. This behavior impairs the expression profiles by homogenizing the gene's activity in the tissue and removing spatial information. 

\subsection{Gene Prediction Benchmark}
  
Figure \ref{fig:general_results}.b shows the performance of all methods for every dataset when trained under our two scenarios (SpaCKLE-completed and raw). It is clear that the prediction performance significantly improves when applying SpaCKLE to every dataset and, in some cases, the best PCC increases to 0.36 points (AHSCC). This result pinpoints the importance of acknowledging missing data for the prediction task and proves the significance of including gene completion in ST pipelines.

Comparing datasets' difficulty, we find that the most challenging dataset to predict was MMBR (PCC=0.16), while 10XGMBSP emerged as the least difficult (PCC=0.74). When inspecting each dataset's characteristics, we observe that, although the organism doesn't seem to have an impact on the difficulty of the task, a larger amount of available genes (due to better quality) and generalizing in an intra-patient fashion typically makes the prediction easier (Supplementary Fig. 1.a and 1.b for tissue type analysis).

\begin{figure}[t] 
    \includegraphics[width=0.99\textwidth]{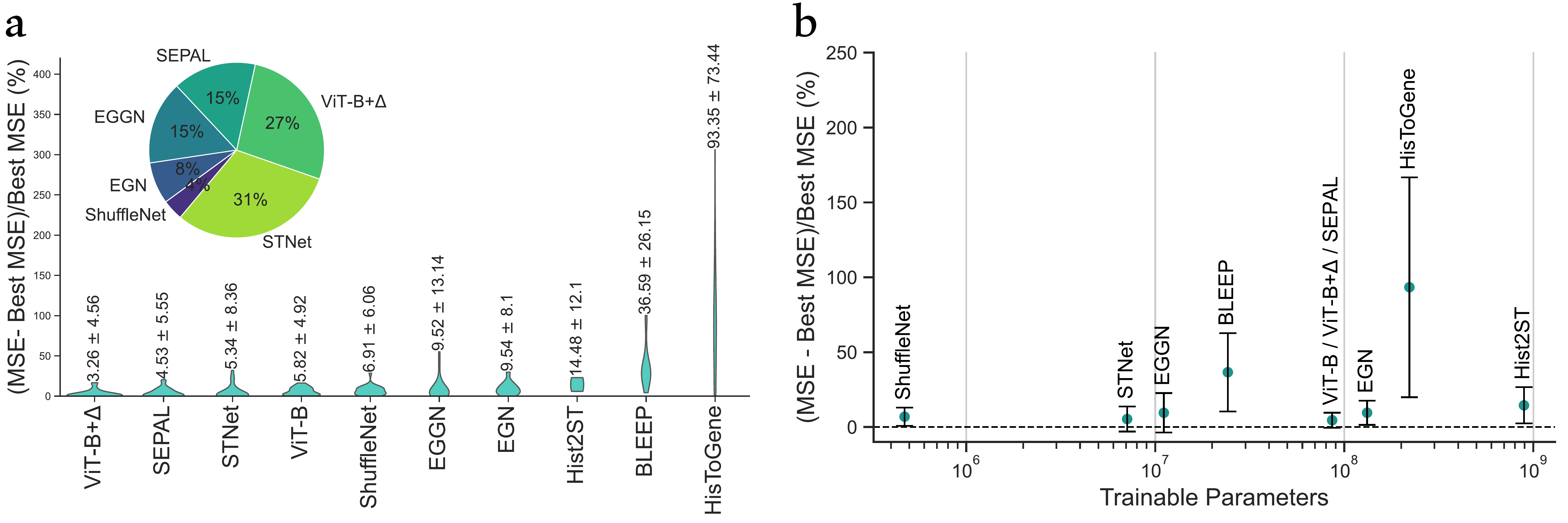}
    \caption{(a) Violin plot: normalized prediction MSE of each model across all datasets within SpaRED, with normalization done against the best MSE obtained on each dataset. The mean and standard deviation of the methods are included at the top of each violin. Pie chart: percentage of datasets within SpaRED for which each model achieves the best prediction MSE. (b) Mean normalized prediction MSE against the number of trainable parameters for each model.} 
    \label{fig:enter-label}
\end{figure}

We display the results of evaluating the 7 state-of-the-art models on SpaRED, as well as the baseline experiments on Fig. \ref{fig:enter-label}.a sorted by best average performance. The normalized MSE metric indicates how close every model's results are to the best performance achieved on each dataset. Results show that ViT-B+$\Delta$ attains the best gene expression predictions on average, despite being one of the most straightforward approaches for the prediction task. Moreover, the pie chart showcases that STNet and ViT-B+$\Delta$ emerge most frequently as the best methods. Interestingly, we notice that SEPAL, which is built on top of ViT-B+$\Delta$, falls behind the latter. This contrast reveals that incorporating local vicinity information does not necessarily improve the outputs and that focusing on predicting the $\Delta$ from the mean expression is already a powerful strategy. Supplementary Table 2 illustrates the statistical differences in MSE performance across all datasets.

In general, most prediction methods exhibit roughly comparable performance. Moreover, our results also indicate that more complex architectures do not necessarily provide superior predictions on our benchmark. This behavior is also supported by Fig. \ref{fig:enter-label}.b, where Hist2ST ranks as the method with the most trainable parameters but performs worse than methods with orders of magnitude fewer parameters. In contrast, ShuffleNet is the method with the fewest parameters and offers a competitive performance. We hypothesize that this counterintuitive trend is caused by the limited scale of publicly available datasets (the biggest SpaRED dataset contains 43,804 spots), probably leading to overfitting in bigger models.

% The previous observations demonstrate a possible performance saturation of the prediction task and a clear need for novel ideas in the field. In addition, these results illustrate the importance of having an appropriate benchmark that allows effective comparison between proposed methods.

\section{Conclusions}

%In this paper, we present SpaRED, an extensive Visium database composed of 26 curated and standardized datasets. We also introduce SpaCKLE, a transformer-based model that successfully completes gene expression values even when the missing data fraction is up to 70\%. In addition, we benchmark 7 state-of-the-art methods in gene expression prediction from histology images, demonstrating that completing training data beforehand using SpaCKLE significantly improves the performance of prediction models. SpaRED emerges as a novel standard point of comparison, and SpaCKLE overcomes the dropout limitations in Spatial Transcriptomics technology. 

% this paper, we present SpaRED, an extensive Visium database composed of 26 curated and standardized datasets. This database emerges as a novel standard point of comparison for the Spatial Transcriptomics research field. We also introduce SpaCKLE, a transformer-based model that successfully completes gene expression values even when the missing data fraction is up to 70\%, thus providing a tool that overcomes the dropout limitations in Spatial Transcriptomics technology. In addition, we benchmark 7 state-of-the-art methods in gene expression prediction from histology images, demonstrating that completing training data beforehand using SpaCKLE significantly improves the performance of prediction models. This work presents powerful tools that we hope will enhance the research development in Spatial Transcriptomics.

In this paper, we present SpaRED, an extensive Visium database composed of 26 curated and standardized datasets, which emerges as a novel standard point of comparison for gene expression prediction from histology images methods. We also introduce SpaCKLE, a transformer-based model that successfully overcomes the dropout limitations, completing gene expression values even when the missing data fraction is up to 70\%. Finally, we benchmark 7 state-of-the-art methods in SpaRED, demonstrating that completing training data beforehand using SpaCKLE significantly improves the performance of prediction models. Consequently, our work represents a significant advancement in the automation of Spatial Transcriptomics and is intended to promote further research in this field.

%Benchmark results reveal that complex models do not significantly outperform simpler ones and highlight potentially enhancing strategies for the prediction task, such as supervising expression changes relative to the mean expression in the training data.

\begin{credits}
\subsubsection{\ackname} Gabriel Mejia and Daniela Vega acknowledge the support of UniAndes-GoogleDeepMind Scholarships 2022 and 2024 respectively.

\subsubsection{\discintname}
The authors have no competing interests to declare that are
relevant to the content of this article.
\end{credits}

\bibliographystyle{splncs04}
\bibliography{Paper-2459}

\end{document}

\title{Supplementary Material for Enhancing Gene Expression Prediction from Histology Images with Spatial Transcriptomics Completion}
\titlerunning{SpaRED}
\author{Gabriel Mejia$^*$
\and
Daniela Ruiz$^*$
\and
Paula Cárdenas
\and
Leonardo Manrique
\and
Daniela Vega
\and
Pablo Arbelaez}
%index{Mejia, Gabriel}
%index{Ruiz, Daniela}
%index{Cárdenas, Paula}
%index{Manrique, Leonardo}
%index{Vega, Daniela}
%index{Arbelaez, Pablo}

\authorrunning{G. Mejia, et al.}
\institute{Center for Research and Formation in Artificial Intelligence \and Universidad de los Andes, Colombia \\
\email{\{gm.mejia,da.ruizl1,p.cardenasg,dl.manrique,\\d.vegaa,pa.arbelaez\}@uniandes.edu.co}}

\maketitle              % typeset the header of the contribution

\begin{figure}[H]
    \centering
    \includegraphics[width=0.84\textwidth]{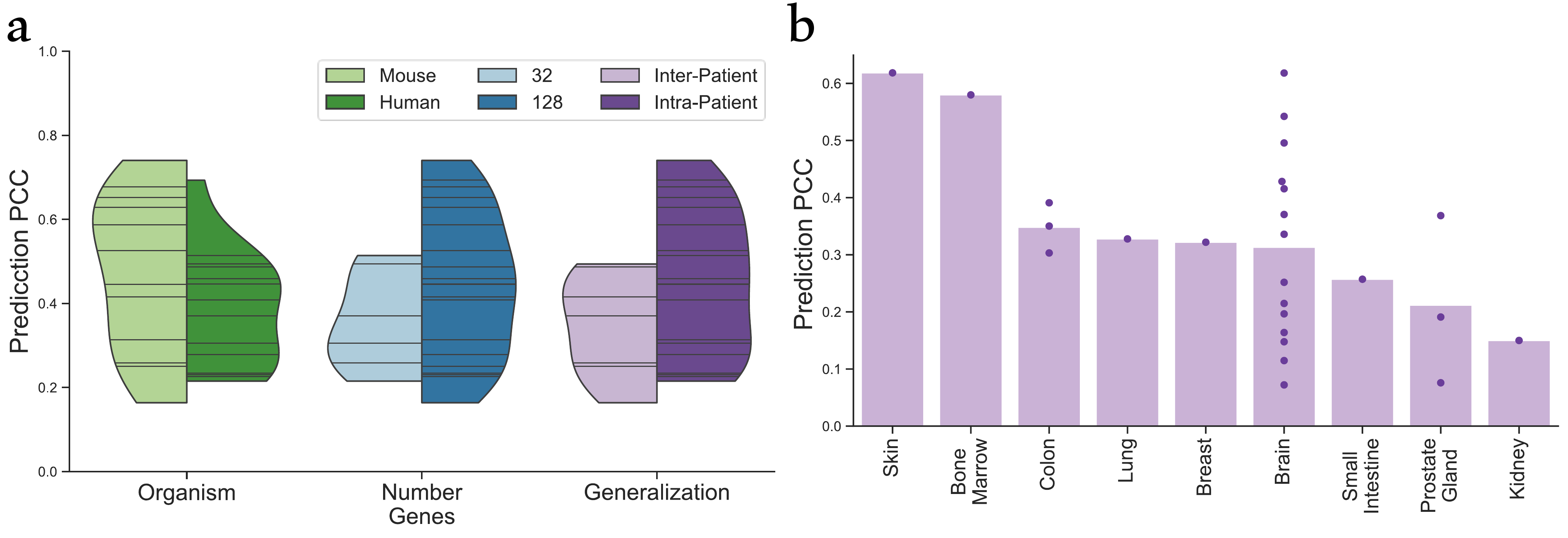}
    \caption{(a) Violin plots illustrate the variation in key characteristics of the datasets, such as organism, number of genes, and generalization task. The data depicted represent the best prediction PCC achieved for each dataset within the SpaRED collection. (b) Bar charts display the prediction performance, measured by PCC, for each type of tissue analyzed in SpaRED. The dots represent the best prediction PCC achieved for each dataset, while the bars indicate the average PCC across each type of tissue.}
    \label{fig:1}
\end{figure}

%\begin{figure}[h]
    %\centering
    %\includegraphics[width=0.5\textwidth]{images/Supplementary_tukey_mat.pdf}
    %\caption{The matrix illustrates the statistically significant differences in MSE among all models across all datasets. The statistical test used was the Tukey Test with a 95\% confidence interval. Cells highlighted in green indicate a significant difference between the corresponding models.}
    %\label{fig:2}
%\end{figure}

%\newpage
%Vicari: 20
%Villacampa: 21
%Mirzazade: 12
%Erikson: 5
%10XGemonics: 18
%Parigi: 15 
%Abalo: 1
%Fan: 6

\begin{table}[t]
\centering
\caption{Detailed overview of the SpaRED datasets, showcasing the generalization task, the number of genes analyzed, the abbreviation for each dataset, the organism studied, the tissue or disease, and the number of slides, patients and spots. Additionally, the table provides the percentages of missing data after processing, labeled as `Processed Missing Data'. The `Corrupt Spots' column indicates the percentage of spots with at least one corrupted value for each dataset. Meanwhile, `Raw Missing Data' represents the percentage of missing data before processing.}

\resizebox{0.99\textwidth}{!}{
\begin{tabular}{c|c|ccccccccc}
\hline
\textbf{\begin{tabular}[c]{@{}c@{}}Generali-\\ zation\end{tabular}}         & \textbf{Genes}        & \textbf{\begin{tabular}[c]{@{}c@{}}Abbreviation/\\ Access\end{tabular}}                                                                                                                                                                                                                                                                  & \textbf{Organism} & \textbf{\begin{tabular}[c]{@{}c@{}}Tissue/\\ Disease\end{tabular}}                                             & \textbf{Slides} & \textbf{Patients} & \textbf{Spots} & \textbf{\begin{tabular}[c]{@{}c@{}}Processed\\ Missing\\ Data\end{tabular}} & \textbf{\begin{tabular}[c]{@{}c@{}}Corrupt \\ Spots\end{tabular}} & \textbf{\begin{tabular}[c]{@{}c@{}}Raw\\ Missing\\ Data\end{tabular}} \\ \hline
\multirow{7}{*}{\begin{tabular}[c]{@{}c@{}}Inter-\\ Patient\end{tabular}}  & \multirow{4}{*}{128}  &     \href{https://figshare.scilifelab.se/articles/dataset/Spatial_Multimodal_Analysis_SMA_-_Spatial_Transcriptomics/22778920}{VMB [20]}                                                                                                                                                                                              & Mouse             & Brain                                                               & 14              & 4                 & 43804         & 28\%                                                              & 100\%                                                             & 89\%                                                         \\
                                &                       &     \href{https://data.mendeley.com/datasets/4w6krnywhn/1}{MMBR [12]}                                                                                                                                                                                                            & Mouse             & Brain                                                               & 8               & 2                 & 34583         & 20\%                                                              & 100\%                                                             & 79\%                                                         \\
                                &                       &     \href{https://data.mendeley.com/datasets/4w6krnywhn/1}{MHPC [12]}                                                                                                                                                                                                                                                                & Human             & Prostate cancer                                                     & 4               & 2                 & 15684         & 21\%                                                              & 100\%                                                             & 79\%                                                         \\
                                &                       &     \href{https://www.ncbi.nlm.nih.gov/geo/query/acc.cgi?acc=GSE169749}{PMI [15]}                                                                                                                                                                                                                                                      & Mouse             & Intestine                                                           & 2               & 2                 & 6234          & 7\%                                                               & 99\%                                                              & 79\%                                                         \\ \cline{2-11} 
                                & \multirow{3}{*}{32}   &     \href{https://data.mendeley.com/datasets/xjtv62ncwr/3}{VLMB [21]}                                                                                                                                                                                               & Mouse             & Brain                                                               & 5               & 2                 & 12202         & 29\%                                                              & 100\%                                                             & 95\%                                                         \\
                                &                       &     \href{https://data.mendeley.com/datasets/4w6krnywhn/1}{MHSI [12]}                                                                                                                                                                                                                 & Human             & Small intestine                                                     & 4               & 2                 & 10474         & 21\%                                                              & 100\%                                                             & 92\%                                                         \\
                                &                       &     \href{https://data.mendeley.com/datasets/4w6krnywhn/1}{MHCP2 [12]}                                                                                                                                                                                                                                                              & Human             & Colon                                                               & 2               & 2                 & 7101          & 24\%                                                              & 100\%                                                             & 97\%                                                         \\ \hline
\multirow{19}{*}{\begin{tabular}[c]{@{}c@{}}Intra-\\ Patient\end{tabular}} & \multirow{16}{*}{128} &     \href{https://data.mendeley.com/datasets/svw96g68dv/4}{EHPCP2[5]}                                                                                                                                                                                                                                                   & Human             & Prostate cancer                                                     & 10              & 1                 & 24465         & 37\%                                                              & 100\%                                                             & 92\%                                                         \\
                                &                       &     \href{https://data.mendeley.com/datasets/svw96g68dv/4}{EHPCP1 [5]}                                                                                                                                                                                                                                                        & Human             & Prostate cancer                                                     & 7               & 1                 & 20987         & 34\%                                                              & 100\%                                                             & 92\%                                                         \\
                                &                       &     \href{https://data.mendeley.com/datasets/4w6krnywhn/1}{MMBP2 [12]}                                                                                                                                                                                                                                                            & Mouse             & Brain                                                               & 4               & 1                 & 17353         & 25\%                                                              & 100\%                                                             & 88\%                                                         \\
                                &                       &     \href{https://data.mendeley.com/datasets/4w6krnywhn/1}{MMBP1 [12]}                                                                                                                                                                                                                                                           & Mouse             & Brain                                                               & 4               & 1                 & 17243         & 11\%                                                              & 100\%                                                             & 70\%                                                         \\
                                &                       &     \href{https://data.mendeley.com/datasets/2bh5fchcv6/1}{AHSCC [1]}                                                                                                                                                                                                                                                           & Human             & \begin{tabular}[c]{@{}c@{}}Squamous \\ cell carcinoma\end{tabular}  & 4               & 1                 & 10374         & 27\%                                                              & 100\%                                                             & 94\%                                                         \\
                                &                       &     \href{https://www.10xgenomics.com/resources/datasets/adult-human-brain-1-cerebral-cortex-unknown-orientation-stains-anti-gfap-anti-nfh-1-standard-1-1-0}{10XGHB [18]}  & Human             & Brain                                                               & 2               & 1                 & 9882          & 23\%                                                              & 100\%                                                             & 89\%                                                         \\
                                &                       &     \href{https://data.mendeley.com/datasets/nrbsxrk9mp/1}{FMBC [6]}                                                                                                                                                                                                                                                               & Mouse             & Brain - Coronal                                                     & 2               & 1                 & 9132          & 24\%                                                              & 100\%                                                             & 80\%                                                         \\
                                &                       &     \href{https://www.10xgenomics.com/resources/datasets/human-breast-cancer-block-a-section-1-1-standard-1-0-0}{10GHBC [18]}                                                                  & Human             & Breast cancer                                                       & 2               & 1                 & 7785          & 12\%                                                              & 100\%                                                             & 79\%                                                         \\
                                &                       &     \href{https://data.mendeley.com/datasets/4w6krnywhn/1}{MMBO [12]}                                                                                                                                                                                                                                                          & Mouse             & Bone                                                                & 4               & 1                 & 7184          & 17\%                                                              & 100\%                                                             & 87\%                                                         \\
                                &                       &     \href{https://www.10xgenomics.com/resources/datasets/mouse-brain-serial-section-2-sagittal-posterior-1-standard-1-1-0}{10XGMBSP [18]}                                        & Mouse             & \begin{tabular}[c]{@{}c@{}}Brain sagittal \\ posterior\end{tabular} & 2               & 1                 & 6644          & 21\%                                                              & 100\%                                                             & 79\%                                                         \\
                                &                       &     \href{https://data.mendeley.com/datasets/4w6krnywhn/1}{MHPBTP1 [12]}                                                                                                                                                                                                                                                        & Human             & \begin{tabular}[c]{@{}c@{}}Pediatric brain \\ tumor\end{tabular}    & 4               & 1                 & 5937          & 21\%                                                              & 100\%                                                             & 73\%                                                         \\
                                &                       &     \href{https://www.10xgenomics.com/resources/datasets/adult-mouse-brain-section-1-coronal-stains-dapi-anti-neu-n-1-standard-1-1-0}{10XGMBC [18]}                                     & Mouse             & Brain coronal                                                       & 2               & 1                 & 5709          & 18\%                                                              & 100\%                                                             & 80\%                                                         \\
                                &                       &     \href{https://www.10xgenomics.com/resources/datasets/mouse-brain-serial-section-1-sagittal-anterior-1-standard-1-0-0}{10XGMBSA [18]}                                         & Mouse             & \begin{tabular}[c]{@{}c@{}}Brain sagittal \\ anterior\end{tabular}  & 2               & 1                 & 5520          & 15\%                                                              & 100\%                                                             & 75\%                                                         \\
                                &                       &     \href{https://data.mendeley.com/datasets/nrbsxrk9mp/1}{FMOB [6]}                                                                                                                                                                                                                                                         & Mouse             & Brain                                                               & 2               & 1                 & 2938          & 10\%                                                              & 100\%                                                             & 67\%                                                         \\
                                &                       &     \href{https://data.mendeley.com/datasets/xjtv62ncwr/2}{VLO [21]}                                                                                                                                                                                                       & Human             & Lung organoids                                                      & 4               & 1                 & 1832          & 26\%                                                              & 100\%                                                             & 90\%                                                         \\
                                &                       &     \href{https://data.mendeley.com/datasets/xjtv62ncwr/1}{VKO [21]}                                                                                                                                                                                                       & Human             & Kidney organoids                                                    & 3               & 1                 & 1355          & 33\%                                                              & 100\%                                                             & 92\%                                                         \\ \cline{2-11} 
                                & \multirow{3}{*}{32}   &     \href{https://figshare.scilifelab.se/articles/dataset/Spatial_Multimodal_Analysis_SMA_-_Spatial_Transcriptomics/22778920}{VHS [20]}                                & Human             & Striatium                                                           & 4               & 1                 & 19033         & 30\%                                                              & 100\%                                                             & 97\%                                                         \\
                                &                       &     \href{https://data.mendeley.com/datasets/4w6krnywhn/1}{MHPBTP2 [12]}                                                                                                                                                                                                                                                      & Human             & \begin{tabular}[c]{@{}c@{}}Pediatric brain \\ tumor\end{tabular}    & 2               & 1                 & 3163          & 30\%                                                              & 100\%                                                             & 97\%                                                         \\
                                &                       &     \href{https://data.mendeley.com/datasets/4w6krnywhn/1}{MHCP1 [12]}                                                                                                                                                                                                                                                             & Human             & Colon                                                               & 2               & 1                 & 2225          & 16\%                                                              & 100\%                                                             & 90\%                                                         \\ \hline
\end{tabular}}
\end{table}

\begin{table}[]
\caption{The matrix illustrates the statistically significant differences in MSE among all models across all datasets. A Dunn test with a 5\% significance level was used to identify these differences. Differences that are statistically significant between models are highlighted in \textbf{bold}.}
\resizebox{0.99\textwidth}{!}{
{\fontsize{7}{10}\selectfont
\begin{tabular}{l|ccccccccc}
\hline
 & \multicolumn{1}{l}{\textbf{ShuffleNet}} & \multicolumn{1}{l}{\textbf{STNet}} & \multicolumn{1}{l}{\textbf{EGN}} & \multicolumn{1}{l}{\textbf{EGGN}} & \multicolumn{1}{l}{\textbf{HisToGene}} & \multicolumn{1}{l}{\textbf{ViT-B}} & \multicolumn{1}{l}{\textbf{VIT-B+$\Delta$}} & \multicolumn{1}{l}{\textbf{SEPAL}} & \multicolumn{1}{l}{\textbf{BLEEP}} \\ \hline
\textbf{ShuffleNet} & - & {1} & 1  & {1} & \textbf{\boldmath 3.01$\times10^{-4}$} & {1} & {0.6028} & {1} & \textbf{0.0023}\\
\textbf{STNet} &  & - & 1 & 1 & \textbf{\boldmath 2.03$\times10^{-8}$} & 1 & 1 & 1 & \textbf{\boldmath 3.02$\times10^{-7}$}\\
\textbf{EGN} &  &  & - & 1 & \textbf{\boldmath 1.21$\times10^{-3}$} & 1 & 0.2465 & 0.861 & \textbf{\boldmath 8.16$\times10^{-3}$} \\
\textbf{EGGN} &  &  &  & - & \textbf{\boldmath 8.75$\times10^{-5}$} & 1 & 1 & 1 & \textbf{\boldmath 7.37$\times10^{-4}$} \\
\textbf{HisToGene} &  &  &  &  & - & \textbf{\boldmath 2.20$\times10^{-5}$} & \textbf{\boldmath 3.17$\times10^{-10}$} & \textbf{\boldmath 6.24$\times10^{-9}$} & 1 \\
\textbf{ViT-B} &  &  &  &  &  & - & 1 & 1 & \textbf{\boldmath 2.09$\times10^{-4}$} \\
\textbf{VIT-B+$\Delta$} &  &  &  &  &  &  & - & 1 & \textbf{\boldmath 5.92$\times10{-9}$} \\
\textbf{SEPAL} &  &  &  &  &  &  &  & - & \textbf{\boldmath 9.91$\times10^{-8}$} \\
\textbf{BLEEP} &  &  &  &  &  &  &  &  & - \\ \hline
\end{tabular}}}
\end{table}